# Thinking Adaptive:
# Towards a Behaviours Virtual Laboratory

**Carlos Gershenson\***
Fundación Arturo Rosenblueth
Facultad de Filosofía y
Letras/UNAM
carlos@jlagunez.iquimica.unam.mx

**Pedro Pablo González\***
\*Instituto de Química/UNAM
Ciudad Universitaria, 04510
México, D. F. México
ppgp@servidor.unam.mx

**José Negrete**
Instituto de Investigaciones
Biomédicas/UNAM
Maestría en Inteligencia
Artificial/UV
jnegrete@mia.uv.mx

## Abstract

In this paper we name some of the advantages of virtual laboratories; and propose that a Behaviours Virtual Laboratory should be useful for both biologists and AI researchers, offering a new perspective for understanding adaptive behaviour. We present our development of a Behaviours Virtual Laboratory, which at this stage is focused in action selection, and show some experiments to illustrate the properties of our proposal, which can be accessed via Internet.

## 1. Introduction

Over the past two decades, research in adaptive behaviour has made clear that properties desirable in autonomous agents (robots, animats, or artificial creatures) can be mapped from properties of animal behaviour (Brooks, 1986; Maes, 1990; Beer, 1990). Behaviour-based systems not only constitute a useful approach for design of autonomous agents, but also an ideal scenery for the development and testing of biological systems theories (Beer, 1993).

Virtual laboratories have been developed in different areas, to reproduce experiments that were made in physical laboratories. Virtual labs are useful for pre practice and post analysis of experiments developed in physical labs, and in some cases they can replace the physical lab itself. Although virtual labs may have many limitations, they have many advantages over physical labs. For example, some physical labs have scarcity of resources (in equipment and staff), limiting the researcher's performance. Virtual labs have relatively low costs, experiments can easily be repeated, and there are no inconveniences in failing experiments. It is desirable that virtual labs exploit the advantages of virtual reality, multimedia, and the Internet. Virtual labs have been developed for different areas, such as physics, electronics, robotics, physiology, chemistry, engineering, economics, and ecology.

We believe that there should be also development of virtual labs in the area of ethology. We name these Behaviour Virtual Laboratories (BVL). This development would benefit both ethology and behaviour-based systems. To ethology, a virtual lab would help reproduce with ease experimental and natural conditions that could take even weeks to develop in a physical lab. For example, some kinds of conditioning in animals take weeks of training, while in a virtual lab, this process may be accelerated, saved, and recovered. For artificial intelligence researchers, a virtual lab would help design and test systems and mechanisms of robots or animats.

A BVL should be capable of achieving the same conditions that are found in an ethology physical laboratory, and even provide better development of the experiments. A Behaviours Virtual Laboratory would be useful to design bottom-up autonomous agents or robots, propose and test animal behaviour theories, reproduce behaviour patterns from experimental data, easily produce lesions in different structures and mechanisms of the animats, amongst other questions. Unlike other types of virtual labs, BVL should be capable of producing unpredictable results, allowing emergent behaviours to arise. With all these properties, a BVL should induce researchers to "think adaptively". This is, to easily show the properties and characteristics of adaptive behaviour, without the need of complex experimentations or heavy research, in an interactive way.

Examples of works related with behaviour virtual laboratories are the Simulated Environment developed by Tyrrell (Tyrrell, 1993), which tests different proposed action selection mechanisms; and Beer's Simulation of Cockroach Locomotion and Escape (Beer, 1993), which allows to lesion different neuronal structures of the insect.

Following these ideas, we are developing a Behaviours Virtual Laboratory, in which animats and simple animat societies can be simulated, having in mind two goals: First, to test and analyse the properties of the Behavioural Columns Architecture (BeCA) action selection mechanism (González, 2000; Gershenson et. al., 2000). Second, to provide a useful tool for biologist and roboticists to experiment with the animal behaviour properties that BeCA is able to simulate. This paper presents the properties of the BVL developed so far, which at this stage is focussed in the action selection problems, rather than in perceptual and motor systems.

In the next section, we present briefly BeCA action selection mechanism. In Section 3 we present the parameters of BeCA that can be modulated in the BVL and the effects they have in the animats' behaviour. Section 4 presents briefly the components of the BVL, and in Section 5 some interesting experiments demonstrating capabilities of the BVL are presented.

## 2. Behavioural Columns Architecture

The Behavioural Columns Architecture is implemented in a distributed blackboard-node architecture (González and Negrete, 1997; Negrete and González, 1998). It consists of two blackboard-nodes: the cognitive node and the motivational node. The cognitive node receives signals from the perceptual system and sends to the motor system which action should be taken, while the motivational node receives signals from the internal medium, and combines the internal and external signals (González et. al., 2000). A diagram of BeCA can be appreciated in Figure 1.

The basic idea of the functioning of BeCA is the following: Each internal behaviour, which is a set of elemental behaviours (or production rules), operates in the different blackboard levels of each node. Each internal behaviour is specified in a particular goal, and all the internal behaviours working together can obtain a solution of the action selection problem. "Behavioural columns" are formed as elemental behaviours of different internal behaviours create solution elements in different blackboard levels that are associated to the same external behaviour.

The structure of an elemental behaviour has three basic elements: a parameters list, a condition part, and an action part. The parameters list specifies which are the condition elements, the action elements, and the coupling strengths related with the elemental behaviour. The condition part generally reads from blackboard levels, with the exceptions of the *exteroceptors* and the *interoceptors*, which receive signals from the perceptual system and the internal medium, respectively. The condition part gives importance to the read signals multiplying them by the coupling strengths (in a similar way as the weights of an artificial neuron). If the condition part is satisfied, then the action part will be executed. The action part will generally inscribe values in a blackboard level, with the exception of the *actuators*, which send signals to the motor system.

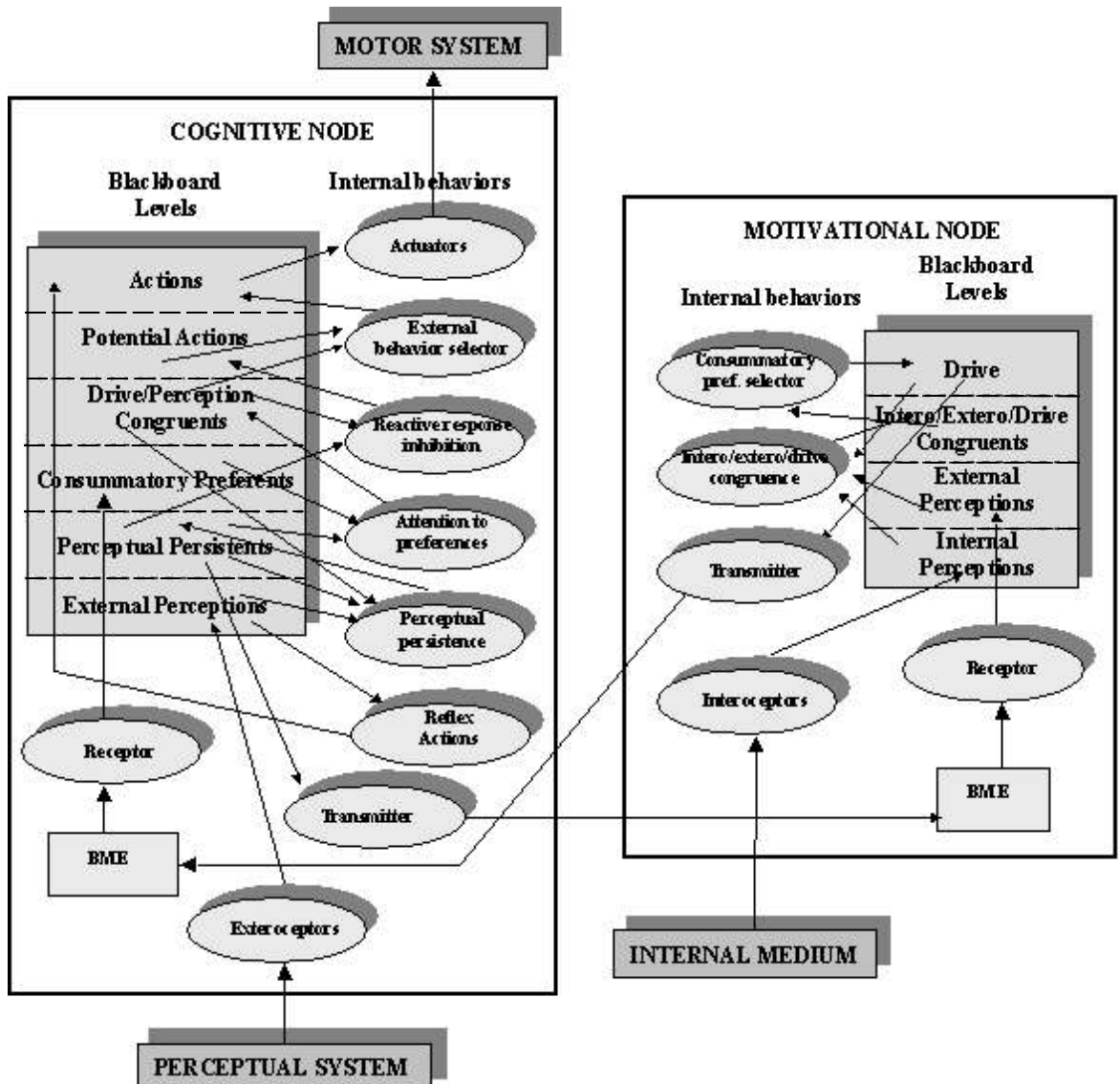

Figure 1. Behavioural Columns Architecture

BeCA presents many properties, such as motivated behaviour, reactive behaviour, reactive response inhibition, pre activation of internal behaviours, and associative learning. It also presents many emergent properties, such as: goal-directed behaviour, non indecision in the action selection, stability in the selection and persistence in the execution of the external behaviours, regulated spontaneity, satiation, changes in responsiveness, and varying attention. At first sight the architecture of BeCA might seem complicated, but for all the properties it presents, it is fairly simple. Other characteristic of BeCA is that it is "context-free". This is, that it can easily be translated into different problem domains, and that it is independent of the environment (because it is independent of the motor and perceptual systems and of the internal medium).

A detailed description of BeCA, its properties, and a comparison with other ASMs can be found in (González, 2000).

## 3. Parameters of BeCA modifiable through the BVL

We have defined parameters that allow to modify and test the properties and behaviours of BeCA. All parameters have values between zero and one.

The main role of the *perceptual persistence* internal behaviour is to represent in the Perceptual Persistents blackboard level the signals that are in the External Perceptions level. Among all the *perceptual persistence* elemental behaviours a competition takes place to determine which elemental behaviours will keep represented for more time the specified signal in its action part. This internal behaviour simulates a short time memory, necessary for the learning processes that take place in BeCA. The time during which this signal will be active in the Perceptual Persistents level will depend of the value of the parameter $\kappa$, which is a decay factor, in expression (1):

$$Atmp_i^T = (1-\kappa)O_i^T + Fa_{ii}^S O_i^S + Fa_{ii}^I O_i^I + \sum_{i \neq j} + Fa_{ij}^T O_j^T \qquad (1)$$

where $O_i^T$ is the strength of the previous signal of the Perceptual Persistents level, $Fa_{ii}^S$ is the coupling strength related to the signal $O_i^S$ of the External Perceptions level, $Fa_{ii}^I$ is the coupling strength of the signal $O_i^I$ of the Drive/Perception Congruents level, and $Fa_{ij}^T$ is the negative coupling strength with which the signal $O_j^T$ laterally inhibits the signal $O_i^T$. The final activation level $A_i^T$ is calculated by hyperbolically converging the activation level $Atmp_i^T$ to a value $Max_i^T$ using expression (2):

$$A_i^T = \begin{cases} \dfrac{-1}{Atmp_i^T + \dfrac{1}{Max_i^T}} + Max_i^T & \text{if } Fa_{ii}^S O_i^S \text{ and } Atmp_i^T > 0 \\ Atmp_i^T & \text{in other cases} \end{cases} \qquad (2)$$

The role of *attention to preferences* internal behaviour is the combination of signals registered on Perceptual Persistents and Consummatory Preferents levels of the cognitive node. The elemental behaviours that compose this internal behaviour function as AND or OR operators, depending of value of $\gamma$ in expression (3):

$$O_i^I = Fa_i^T O_i^T \left( \gamma + \phi \dfrac{\sum_j Fa_j O_j^C}{O_i^T} \right) \qquad (3)$$

where $O_i^I$ is the value to be inscribed in the Drive/Perception Congruents level; $O_i^T$ is the signal from the Perceptual Persistents level and $Fa_i^T$ its corresponding coupling strength; $O_j^C$ is the signal from the Consummatory Preferents level and $Fa_j^C$ its corresponding coupling strength; and $\gamma$ and $\phi$ modulate the reactivity degree in the observed behaviour of the agent.

For a value of $\phi$ equal to one, the integral value of the signals from the motivational node is taken, making the external behaviour motivated. As $\phi$ decreases, less importance is given to the signal from the motivational node, making the external behaviour less motivated. If $\phi$ is equal to zero, there will be no flow of signals from the motivational node, and the agent will not have any knowledge of its internal needs. Therefore, with $\phi$ we can produce a kind of lesion in the mechanism.

For a value of $\gamma$ greater than zero, greater importance is given to the external stimuli, represented by the signals of the Perceptual Persistents level, than to the signals from the motivational node, found in the Consummatory Preferents level. This makes that yet in the absence of motivation for an external behaviour, this might be executed reactively.

The role of the *intero/extero/drive congruence* internal behaviour is to combine signals from the Internal Perceptions, External Perceptions, and Drive levels of the motivational node blackboard. The model for combination of internal and external signals is given by expression (4):

$$A_i^C = Fa_i^E O_i^E \left( \alpha + \sum_j Fa_{ij}^S O_j^S \right) + Fa_i^D O_i^D \qquad (4)$$

where $A_i^C$ is the new activation level correspondent to the Intero/Extero/Drive Congruents level of the motivational node; $O_i^E$ is the signal from the Internal Perceptions level and $Fa_i^E$ its coupling strength; $O_j^S$ is the signal from the External Perceptions level and $Fa_{ij}^S$ its coupling strength; $O_i^D$ is the signal from the Drive level and $Fa_i^D$ its coupling strength; and $\alpha$ regulates the combination of the internal and external signals.

This combination model is discussed in detail in (González et. al., 2000).

For a value of α equal to zero, the internal signal and external signals interact in a multiplicative way. This makes that if one of the signals (internal or external) is very small, it lessens the importance of the other signal. In this way, external signals that contribute to weak motivations, will make the correspondent external behaviour to have little chance to be selected. The same happens with small external signals for strong motivations.

If we consider a value of α greater than zero, then the internal state will have more importance than the external signal. In this way, external signals that contribute to strong motivations, will make the correspondent external behaviour to have strong chance to be selected, even in the total absence of external signals. This makes the external behaviour to be motivated.

BeCA models associative learning (primary and secondary classical conditionings) adjusting the coupling strengths of certain internal behaviours. The rule that is used for this adjustment is given by expression (5):

$$Fa_{ij}(t+1) = \begin{cases} (1-\beta)Fa_{ij}(t) + \lambda\left(f\left(O_j^{in}(t)\right)f\left(O_i^{out}(t)\right)\right) \\ \qquad\qquad\qquad if\ f\left(O_j^{in}(t)\right) > 0 \\ (1-\mu)Fa_{ij}(t) \qquad in\ other\ case \end{cases} \quad (5)$$

where $O_j^{in}$ is the value of the signal specified in the condition of the elemental behaviour; $O_i^{out}$ is the value of the signal specified in the action of the elemental behaviour; and β, λ, and μ are parameters to modulate the learning processes.

β determines the proportion that is kept from the previous coupling strength in a conditioning stage.

λ regulates the speed of the conditioning. For greater values of λ, fewer presentations will be needed to achieve a conditioning.

μ determines the speed of the extinction of the conditioning. If μ is equal to zero, there will be no extinction of the conditioning.

## 4. The Behaviours Virtual Laboratory

Our approximation to a BVL at this stage of development is focussed to the action selection. We built a virtual environment where the user can create different kinds of external stimuli (such as food sources, water sources, walls, etc.) and animats (predators or preys) through a friendly interface. We developed experiments using one animat to test the properties of BeCA (Gershenson et. al., 2000; González, 2000), and now it supports several animats. At this stage, the BVL can't be expanded by the user (in defining new behaviours and stimuli), so it does not proportionate all the characteristics desirable in a BVL. This BVL can be accessed and/or downloaded via Internet in the URL: http://132.248.11.4/~carlos/asia/bvl

### 4.1. The animats

We developed animats of two kinds: predators and preys. Predators chase and eat preys, and preys run away from the predators. Our initial intention was not to reproduce the behaviour of specific species of animals, but to model general properties found in animal behaviour.

The internal structure of each animat can be described in terms of four basic components: the perceptual system, the internal medium, the action selection mechanism (BeCA), and the motor system.

The perceptual system first registers stimuli that are in the perceptual region ($R_p$) found in the plane (z, x) of the space (x, y, z) defined by the half-circle of expression (6):

$$R_p = \begin{cases} \left(\left(x > x_a + \tan\left(\left(\theta + \frac{\pi}{2}\right)(z - z_a)\right)\right) \\ \cap\left((z-z_a)^2 + (x-x_a)^2 < r_p^2\right)\right) & if\ 0 < \theta \le \pi \\ \left(\left(x < x_a + \tan\left(\left(\theta + \frac{\pi}{2}\right)(z - z_a)\right)\right) \\ \cap\left((z-z_a)^2 + (x-x_a)^2 < r_p^2\right)\right) & if\ \pi < \theta \le 2\pi \end{cases} \quad (6)$$

where ($z_a$, $x_a$) is the position of the animat, θ its orientation in radians, and $r_p$ is the radius of the half-circle. After this, the perceptual system eliminates the stimuli that are found behind obstacles, as shown in Figure 2, determining the "perceived scenario". The stimuli found in the perceived scenario are pondered as a ratio between the magnitude of the stimulus and its distance from the animat. If a stimulus leaves the perceived scenario, then the pondered value (Fe) decreases in terms of the parameter ℵ, as shown in expression (7):

$$Fe(t+1) = Fe(t) - \aleph \quad (7)$$

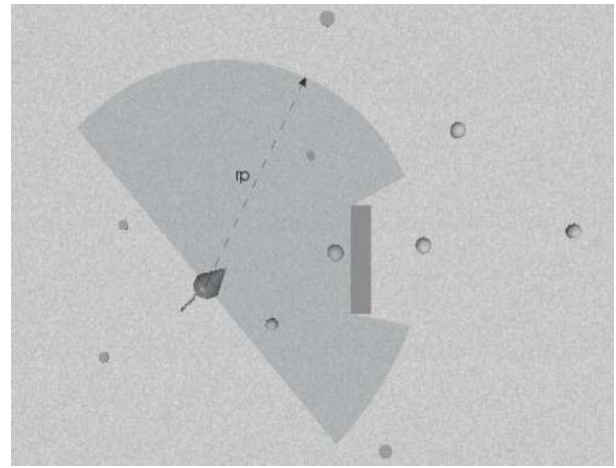

Figure 2. Perceived scenario of an animat.

Expression (7) simulates a short-medium time memory. The "remembered" stimuli conform the animat's "remembered scenario". All the stimuli found in the perceived and remembered scenarios are registered in BeCA by the *exteroceptors*.

The animat's movement is commanded by angular steps α and β, with a centre in the extremes of the diameter of the projection of the sphere of the animat in the plane (z, x), as shown in Figure 3.

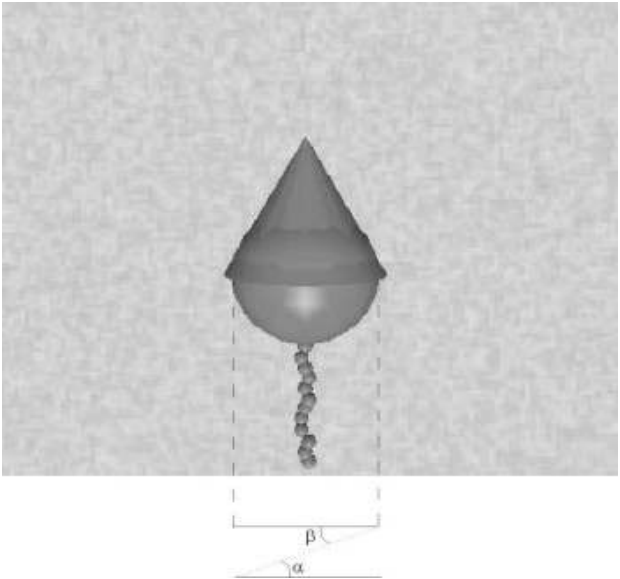

Figure 3. Angular steps of the animat.

The motor system receives signals from BeCA through the *actuators*, and can execute the next behaviours: wander, explore, approach (to different stimuli), eat, drink, rest, runaway (from different stimuli), and the reflex behaviour avoid obstacle.

The internal medium is defined by a set of variables which can take values between zero and one, representing strength, lucidity, safety, fatigue, thirst, and hunger. The size of the angular steps of the animat is proportional to his strength, while its radius of perception is proportional to his lucidity. The safety value does not change in time; but fatigue, thirst and hunger are increased in time (or decreased if a proper consummatory behaviour is executed). The internal medium of the animat is perceived by the *interoceptors* of BeCA.

## 4.2. The Virtual Environment

The virtual environment is defined by a plane (z, x), limited by a frame, of a space (x, y, z). In the area defined by this frame different objects can be created. This objects represent the external stimuli food (green spheres), water (blue circles), grass (texturized green circles), fixed obstacles (brown parallelepipeds), blobs (black ellipsoids), and other kinds of stimuli that initially have no specific meaning for the entity (red and yellow circles). The frame that delimits the plane (z, x) is also considered as a fixed obstacle. The animats perceive these stimuli, and act upon them. Figure 4 shows an aerial view of the simulated environment.

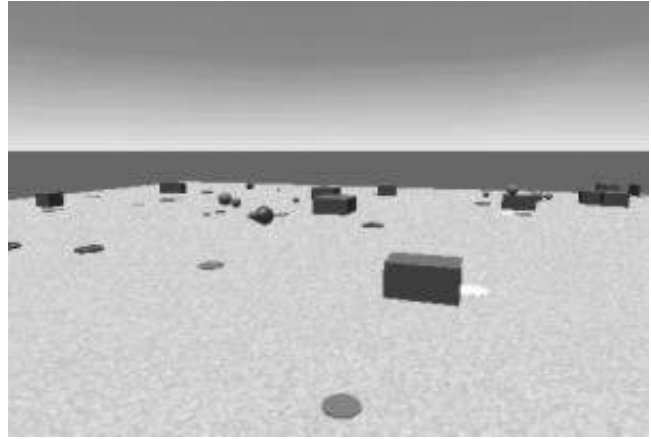

Figure 4. Virtual environment

## 4.3. The Interface

The interface of the BVL allows to perform a wide variety of simulations and experiments. It consists of one window containing the general controls of the BVL and one window for each animat created in the BVL, as the ones shown in Figures 5 and 6.

The general controls allow the user to save, load, and reset animats, environments, and simulations. Animats are saved with all their properties (internal states, learning states, parameters, and attributes). Simulations handle animats and environments as one. This allows to save initial, partial, or final states of experiments easily. In this window, the user can add and remove external stimuli, randomly or with specific positions and magnitudes; pause and resume the simulation; and set a delay for each simulation interval in milliseconds.

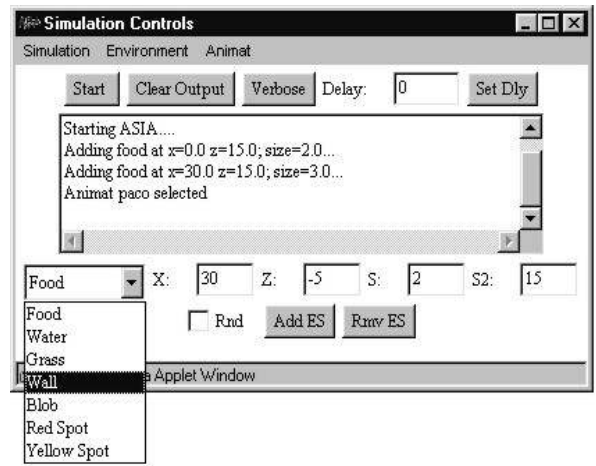

Figure 5. General controls window.

In the animat controls, the user can set the name of the animat, its position and orientation, its radius of perception, and its type (predator or prey). The animat also can be set as immortal. The internal states of the animat are adjusted and shown in the same display, which is a set of scrollbars. The animat can leave a trail, which colour can be also selected to: a specific colour, the animat's colour, or the RGB colour of the magnitudes of fatigue, thirst, and hunger mapped to red, blue, and green, respectively. The parameters α, β, γ, κ, λ, and μ of BeCA; and ℵ of the perceptual system, can be modified through this interface.

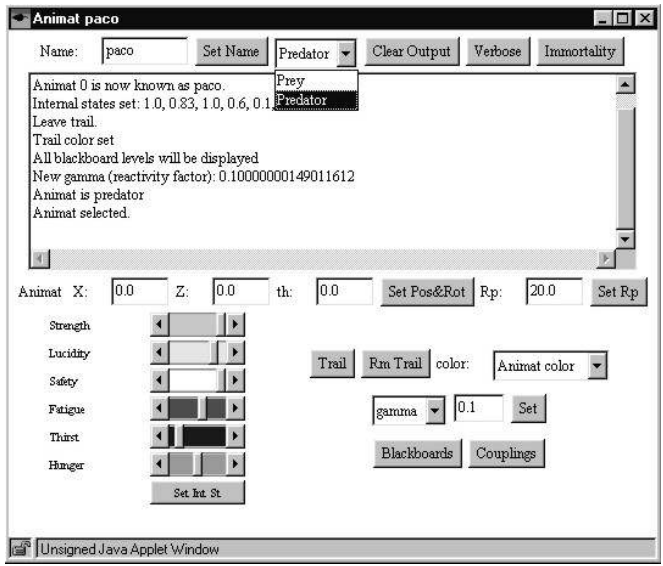

Figure 6. Animat controls window.

Both controls have a display to inform of the states of the simulation, environment, or animat. The animat controls can display the actual state of the blackboard levels, and of the coupling strengths involved in the learning processes.

All the presented properties of the BVL allow a wide variety of situations in order to produce experiments and simulations.

## 5. Experiments

In this section we'll describe two series of experiments that try to show some of the capabilities of BeCA and the BVL. In the first series of experiments, we'll observe what changes in the external behaviour of an animat are produced by changing the parameters that modulate motivation and reactiveness. In the second series, we'll show the role of short-time memories in the learning processes; and how these affect the chances of survival of animats.

### *5.1 Motivated and reactive behaviours*

In these experiments, we modified the values of the parameters α, γ, and ϕ; in order to observe how motivated or reactive is an animat behaviour depending on these parameters.

We used for all experiments an initial state as the one shown in Figure 7. The animat has little fatigue, much thirst, and some hunger. There are food sources near him, but the water sources are distant and the animat can't perceive them at this stage. The BVL's interface allows to easily load this initial state for each experiment.

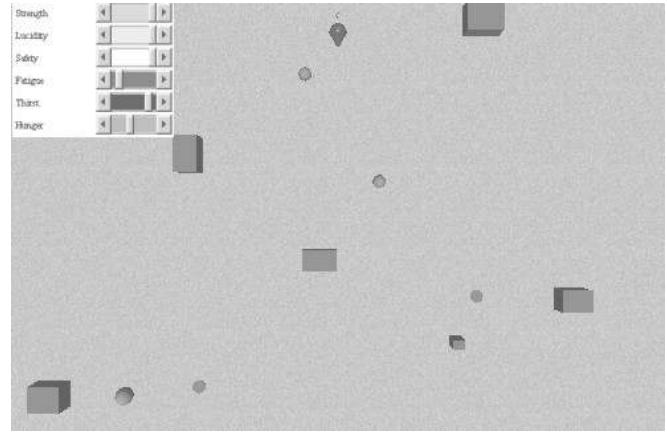

Figure 7. Initial state of the experiments of Section 5.1

First, we took values of α=0.8, γ=0.0, and ϕ=1.0. These are the default values used in the BVL. The behaviours executed by the animat can be appreciated in Figure 8. Since the animat had some hunger, he approached first to the food source nearest to him, and ate until the hunger was reduced (but not totally). Then he began to explore in search of water, in order to satisfy his thirst. Once he perceived it, he approached to it, and began to drink. These behaviours were motivated by the internal states of the animat.

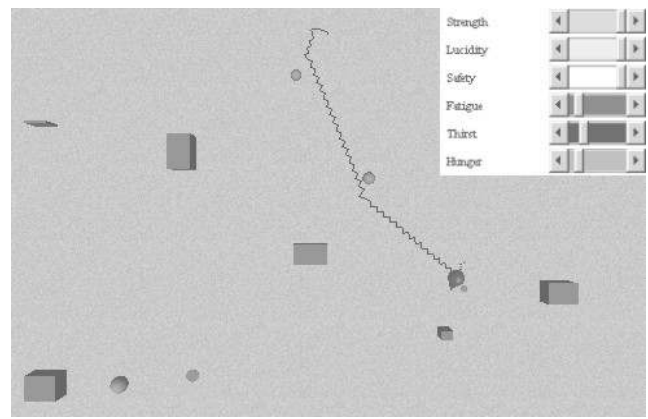

Figure 8. Behaviours executed with α=0.8, γ=0.0, and ϕ=1.0.

For the next experiment, we took values of α=0.0, γ=0.0, and ϕ=1.0. Since the internal and external signals are being combined multiplicatively (as can be seen in expression (4)), the animat will need to perceive the water before the respective behavioural column may win the competence in the motivational node. So, since the animat was perceiving food and was hungry, he approached it and ate until his hunger was satiated. After this, he wandered until he finally perceived the water source, approached it, and satisfied his thirst by drinking it; as shown in Figure 9. These behaviours are less motivated, since the animat can't execute the explore behaviour, and any small internal need may fire its behaviour only if a correspondent external stimulus is perceived. This also affects the animat's survival performance, because he will need more time to find a stimulus to satiate a need than for higher values of α (González et. al., 2000).

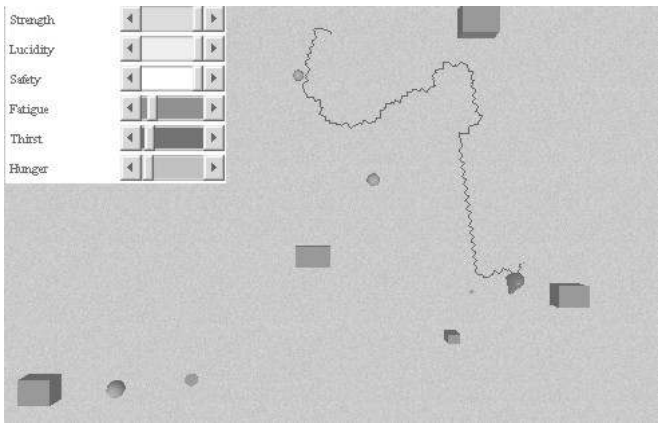

Figure 9. Behaviours executed with α=0.0, γ=0.0, and ϕ=1.0.

For the following experiment, we used values of α=0.8, γ=0.0, and ϕ=0.0. Since ϕ=0 causes that no signal from the motivational node reaches the cognitive node, the animat has no awareness of his internal needs. Because of this, he will wander independently of his perceptions or needs (except for reflex behaviours); as shown in Figure 10, until his death.

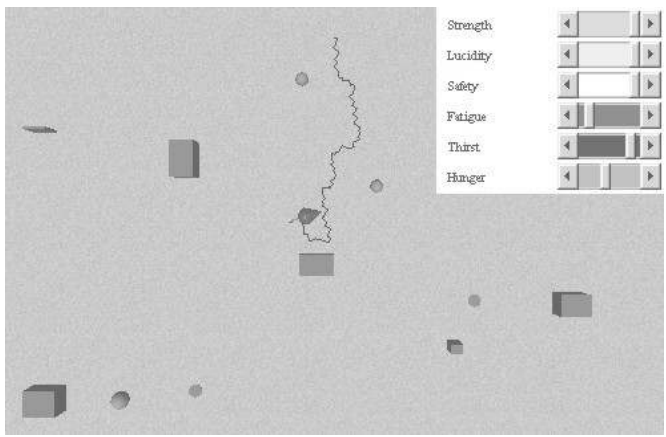

Figure 10. Behaviours executed with α=0.8, γ=0.0, and ϕ=0.0.

For values of α=0.0, γ=0.0, and ϕ=0.0 we had similar results. If there is no flow of signals from the motivational node, the value of α will not affect the external behaviour of the animat.

Next, we used values of α=0.8, γ=0.1, and ϕ=1.0. γ greater than zero in expression (3) allows a behaviour to be reactively executed, even in the absence of an internal need for it. Figure 11 shows the behaviours that the animat executed: First, he perceived food, approached it, and ate it completely, even when he had no more hunger. Then, he tried to approach another food source, until he perceived the water source, approached it, and drank it completely. Since he had little fatigue, he began to explore in search of a grass where to rest.

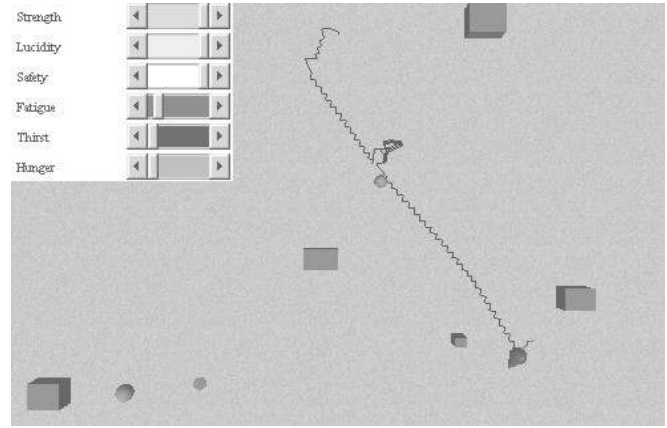

Figure 11. Behaviours executed with α=0.8, γ=0.1, and ϕ=1.0.

With values of α=0.0, γ=0.1, and ϕ=1.0 we had the same results.

Figure 12 shows the last experiment, where we first took values of α=0.8, γ=0.1, and ϕ=0.0. In this case, the animat has no awareness of his internal state, but he acts reactively since γ is greater than zero. So he had a similar behaviour as the one shown in the previous experiment, only that after drinking all the water, he wanders because of his unawareness. In this situation, the animat would only survive if he would find casually external stimuli for the needs he is having in a precise moment. If he would not run into an appropriate external stimulus, he would die unknowingly.

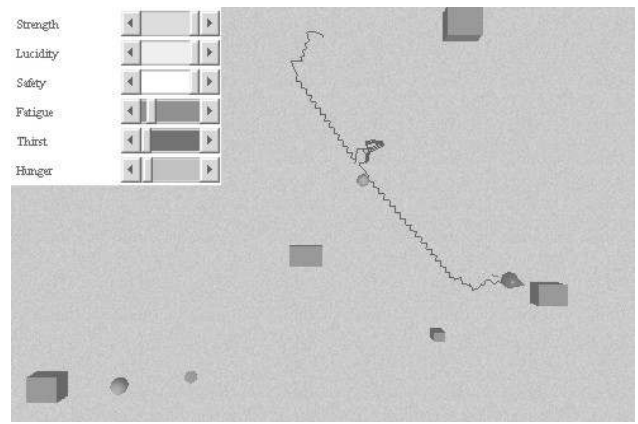

Figure 12. Behaviours executed with α=0.8, γ=0.1, and ϕ=0.0.

With values of α=0.0, γ=0.1, and φ=0.0 we had similar results.

## 5.2. The Role of Short-time Memory in the Learning Processes

In these experiments, we lesioned the short time memories: the one simulated by the *perceptual persistence* internal behaviour, and the one simulated by the perceptual system. These were altered by modifying the parameters κ and ℵ, respectively.

First, we used a prey animat A, with the default values κ=0.25 and ℵ=0.1. He leaves a black trail, while he is approaching a closed zone where a satiated predator wanders, leaving a white trail. We put a neutral stimulus (a red spot) near the entrance to the zone where the predator roams. Since we wanted the animat to associate the red spots with predators, we used a value of λ=1.0. With these values, the animats executed the behaviours shown in Figure 13: Prey animat A was thirsty, and exploring in search of water. He perceived the red spot before he perceived the predator, and he ran away; but he already associated the red spot with the predator, because of the high value of λ. We saved this conditioned prey animat A for further experimentation.

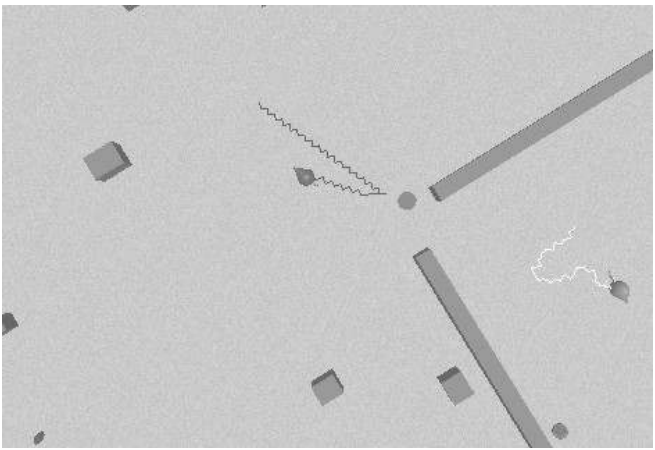

Figure 13. Prey animat running away from wandering predator, after perceiving a neutral stimulus.

Next, we used a prey animat B, in the same conditions as prey animat A; only that we set the values κ=1.0 and ℵ=1.0. This means that a signal won't reverberate in the Perceptual Persistence level, due to the value of κ in expression (1); and that the remembered scenario will be "forgotten" almost immediately, due to the value of ℵ in expression (7). We observed the same behaviours as in the previous experiment in the prey animat A. Only that, even for a value of λ=1.0, the red spot couldn't be associated with the predator. We saved prey animat B for further experimentation.

For the next experiment, we loaded two prey animats A in the environment, and set them in a position so that they would approach the zone where now two hungry predators roamed. Since we successfully conditioned prey animats A, when the preys perceived the red spot, they ran away, without the need of perceiving a predator (and allowing him to perceive them). These behaviours can be appreciated in Figure 14. Prey animats are leaving a black trail, while predators are leaving a white one.

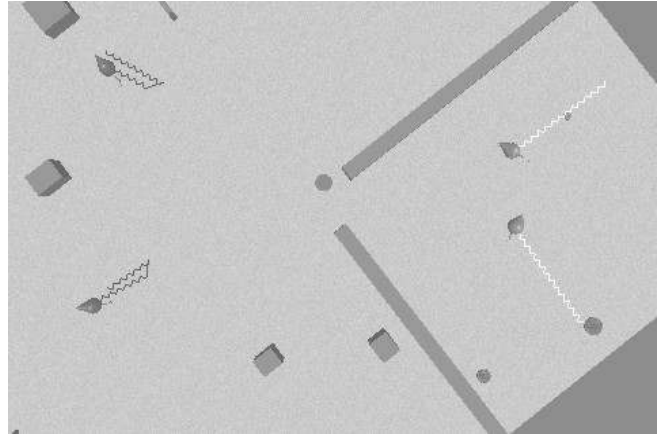

Figure 14. Prey animats A running away from a red spot previously associated with presence of predators.

For our last experiment, we loaded prey animat B, in a situation such that he would approach the zone with two hungry predators. Since the red spot acquired no meaning for him, he naively entered into the predators' zone, which now were hungry and exploring in search of preys. As soon as these perceived him, they began to chase him. Since the animat had no memory, as soon as he began to run away from a predator, he forgot that he was being chased. He then perceived the other predator, and ran away in the other direction. Finally the predators easily caught him and satiated their hunger, as shown in Figure 15.

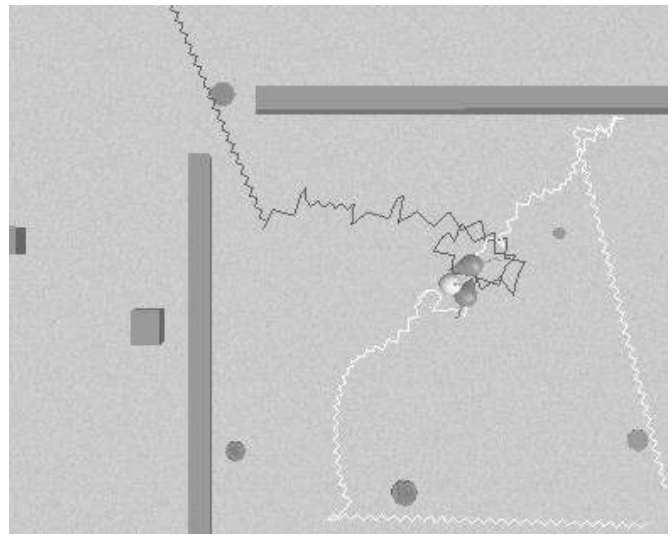

Figure 15. Prey animat B chased and captured by predators.

## Conclusions

We presented our approximation to a Behaviours Virtual Laboratory. It still lacks many properties desirable in a Behaviours Virtual Laboratory, but at this stage it would be already useful for biologists and AI researchers. We propose that a BVL would be helpful not only in experimentation and design, but also an instrument to think with the understanding of adaptive behaviour.

We believe that a BVL like the one we are developing would also help to divulge in the scientific community models or experiments that in many cases are developed in simulations that only the developers can access to. This would also produce valuable feedback among researchers of different areas.

Our future work is directed to evolving the BVL with feedback of researchers in the areas of ethology, artificial intelligence, life sciences, and complex systems. As a short term goal, we intend that parameters of BeCA are learned depending on the animat's experience. Also, we intend to implement instrumental conditioning and learning at a cognitive level in BeCA. As a medium term goal, we intend to study the role of emotions (Gershenson, 1999) in the action selection process; and also develop communication methods between the animats, to simulate more complex social behaviours. As a long term goal, we intend to include evolutive processes in a complex animat society. An Evolution Virtual Laboratory would be extremely useful for proposing and testing theories of evolution of life, societies, and cultures.